\documentclass[11pt]{article}
\usepackage{eacl2017}
\usepackage{times,booktabs}
\usepackage{url}
\usepackage{latexsym}
\usepackage{graphicx}
\usepackage{tabularx}
\eaclfinalcopy 
\def\eaclpaperid{485} 

\newcommand\MTL{\cite{Klerke:ea:16,Soegaard:Goldberg:16,Bollman:Soegaard:16,Plank:16,Braud:ea:16,Alonso:Plank:17}}

\title{Identifying beneficial task relations for multi-task learning\\ in deep neural networks}

\author{
  Joachim Bingel \\
  Department of Computer Science \\
  University of Copenhagen \\
  {\tt bingel@di.ku.dk} \\\And
  Anders S{\o}gaard\thanks{\hspace{.3em} Both authors contributed to the paper in equal parts.} \\
  Department of Computer Science \\
  University of Copenhagen \\
  {\tt soegaard@di.ku.dk}\\}

\date{}

\begin{document}
\maketitle
\begin{abstract}
Multi-task learning (MTL) in deep neural networks for NLP has recently received increasing interest due to some compelling benefits, including its potential to efficiently regularize models and to reduce the need for labeled data. While it has brought significant improvements in a number of NLP tasks, mixed results have been reported, and little is known about the conditions under which MTL leads to gains in NLP. This paper sheds light on the specific task relations that can lead to gains from MTL models over single-task setups.
\end{abstract}

\section{Introduction}

Multi-task learning is receiving increasing interest in both academia and industry, with the potential to reduce the need for labeled data, and to enable the induction of more robust models. The main driver has been empirical results pushing state of the art in various tasks, but preliminary theoretical findings guarantee that multi-task learning works under various conditions. Some approaches to multi-task learning are, for example, known to work when the tasks share optimal hypothesis classes \cite{Baxter:00} or are drawn from related sample generating distributions \cite{Ben-David:ea:03}.

In NLP, multi-task learning typically involves very heterogeneous tasks. However, while great improvements have been reported \cite{Luong:ea:16,Klerke:ea:16}, results are also often mixed  \cite{Collobert:Weston:08,Soegaard:Goldberg:16,Alonso:Plank:17}, and theoretical guarantees no longer apply. The question {\em what task relations guarantee gains or make gains likely in NLP}~remains open. 

\paragraph{Contributions} This paper presents a systematic study of {\em when}~and {\em why} MTL works in the context of sequence labeling with deep recurrent neural networks. We follow previous work \MTL~in studying the set-up where hyperparameters from the single task architectures are reused in the multi-task set-up (no additional tuning), which makes predicting gains feasible. Running MTL experiments on 90 task configurations and comparing their performance to single-task setups, we identify data characteristics and patterns in single-task learning  that predict task synergies in deep neural networks. Both the LSTM code used for our single-task  and multi-task models, as well as the script we used for the analysis of these, are  available at \url{github.com/jbingel/eacl2017_mtl}.

\section{Related work} 
In the context of structured prediction in NLP, there has been very little work on the conditions under which MTL works. \newcite{Luong:ea:16}~suggest that it is important that the auxiliary data does not outsize the target data, while \newcite{Hovy:ea:17} suggest that multi-task learning is particularly effective when we only have access to small amounts of target data. \newcite{Alonso:Plank:17} present a study on different task combinations with dedicated main and auxiliary tasks. Their findings suggest, among others, that success depends on how uniformly the auxiliary task labels are distributed. 

\newcite{Mou:2016:transferable} investigate multi-task learning and its relation to transfer learning, and under which conditions these work between a set of sentence classification tasks. Their main finding with respect to multi-task learning is that success depends largely on ``how similar in semantics the source and target datasets are'', and that it generally bears close resemblance to transfer learning in the effect it has on model performance.

\section{Multi-task Learning}

While there are many approaches to multi-task learning, hard parameter sharing in deep neural networks \cite{Caruana:93} has become extremely popular in recent years. Its greatest advantages over other methods include  (i) that it is known to be an efficient regularizer, theoretically \cite{Baxter:00}, as well as in practice \cite{Soegaard:Goldberg:16}; and (ii) that it is easy to implement. 

The basic idea in hard parameter sharing in deep neural networks is that the different tasks share some of the hidden layers, such that these learn a joint representation for multiple tasks. Another conceptualization is to think of this as regularizing our target model by doing model interpolation with auxiliary models in a dynamic fashion. 

Multi-task linear models have typically been presented as matrix regularizers. The parameters of each task-specific model makes up a row in a matrix, and multi-task learning is enforced by defining a joint regularization term over this matrix. One such approach would be to define the joint loss as the sum of losses and the sum of the singular values of the matrix. The most common approach is to regularize learning by the sum of the distances of the task-specific models to the model mean. This is called mean-constrained learning. Hard parameter sharing can be seen as a very crude form of mean-constrained learning, in which parts of all models (typically the hidden layers) are enforced to be identical to the mean. 

\begin{table*}[t!]
\begin{center}
\begin{tabular}{l|rrrrrrr|c}
Task & Size & \# Labels & Tok/typ & \%OOV & $H(y)$& $||X||_F$& JSD & $F_1$ \\
\hline
\textsc{ccg} &  39,604&   1,285&  23.08& 1.13& 3.28& 981.3& 0.41 & 86.1\\
\textsc{chu} &   8,936&      22&  12.01& 1.35& 1.84& 466.4& 0.47 & 93.9\\
\textsc{com} &   9,600&       2&   9.47& 0.99& 0.47& 519.3& 0.44 & 51.9\\
\textsc{fnt} &   3,711&       2&   8.44& 1.79& 0.51& 286.8& 0.30 & 58.0\\
\textsc{pos} &   1,002&      12&   3.24&14.15& 2.27& 116.9& 0.24 & 82.6\\
\textsc{hyp} &   2,000&       2&   6.14& 2.14& 0.47& 269.3& 0.48 & 39.3\\
\textsc{key} &   2,398&       2&   9.10& 4.46& 0.61& 289.1& 0.39 & 64.5\\
\textsc{mwe} &   3,312&       3&   9.07& 0.73& 0.53& 217.3& 0.18 & 43.3\\
\textsc{sem} &  15,465&      73&  11.16& 4.72& 2.19& 614.6& 0.35 & 70.8\\
\textsc{str} &   3,312&     118&   9.07& 0.73& 2.43& 217.3& 0.26 & 61.5\\
\end{tabular}
\caption{Dataset characteristics for the individual tasks as defined in Table \ref{datafeats}, as well as single-task model performance on test data (micro-averaged $F_1$).
} \label{table:datasets}

\end{center}
\end{table*}

Since we are only forcing parts of the models to be identical, each task-specific model is still left with wiggle room to model heterogeneous tasks, but the expressivity is very limited, as evidenced by the inability of such networks to fit random noise \cite{Soegaard:Goldberg:16}. 

\subsection{Models} Recent work on multi-task learning of NLP models has focused on sequence labeling with recurrent neural networks \MTL, although sequence-to-sequence models have been shown to profit from MTL as well \cite{Luong:ea:16}.
Our multi-task learning architecture is similar to the former, with a bi-directional LSTM as a single hidden layer of 100 dimensions that is shared across all tasks. The inputs to this hidden layer are 100-dimensional word vectors that are initialized with pretrained GloVe embeddings, but updated during training. The embedding parameters are also shared. The model then generates predictions from the bi-LSTM through task-specific dense projections. Our model is symmetric in the sense that it does not distinguish between main and auxiliary tasks.

In our MTL setup, a training step consists of uniformly drawing a training task, then sampling a random batch of 32 examples from the task's training data. Every training step thus works on exactly one task, and optimizes the task-specific projection and the shared parameters using Adadelta. As already mentioned, we keep hyper-parameters fixed across single-task and multi-task settings, making our results only applicable to the scenario where one wants to know whether MTL works in the current parameter setting \cite{Collobert:Weston:08,Klerke:ea:16,Soegaard:Goldberg:16,Bollman:Soegaard:16,Plank:16,Braud:ea:16,Alonso:Plank:17}.

\subsection{Tasks}
In our experiments below, we consider the following ten NLP tasks, with one dataset for each task. Characteristics of the datasets that we use are summarized in Table \ref{table:datasets}.
\begin{enumerate}
\item \textbf{CCG Tagging} ({\sc ccg}) is a sequence tagging problem that assigns a logical type to every token. We use the standard splits for CCG super-tagging from the CCGBank \cite{Hockenmaier:2007:CCC}. 

\item {\textbf{Chunking}} ({\sc chu}) identifies continuous spans of tokens that form syntactic units such as noun phrases or verb phrases. 
We use the standard splits for syntactic chunking from the English Penn Treebank \cite{Marcus:ea:93}. \item  {\textbf{Sentence Compression}} ({\sc com}) We use the publicly available subset of the Google Compression dataset \cite{filippova:13}, which has token-level annotations of word deletions. 
\item  {\textbf{Semantic frames}} ({\sc fnt}) We use FrameNet~1.5 for jointly predicting target words that trigger frames, and deciding on the correct frame in context. 
\item {\textbf{POS tagging}} ({\sc pos}) 
We use a dataset of tweets annotated for Universal part-of-speech tags \cite{Petrov:ea:11}. 
\item  {\textbf{Hyperlink Prediction}} ({\sc hyp}) 
We use the hypertext corpus from \newcite{spitkovsky2010} and predict what sequences of words have been bracketed with hyperlinks. 
\item  {\textbf{Keyphrase Detection}} ({\sc key}) This task amounts to detecting keyphrases in scientific publications. We use the SemEval 2017 Task 10 dataset. 
\item  {\textbf{MWE Detection}} ({\sc mwe}) We use the Streusle corpus \cite{schneider2015corpus} to learn to identify multi-word expressions (\textit{on my own, cope with}).
\item  {\textbf{Super-sense tagging}} ({\sc sem}) We use the standard splits for the Semcor dataset, predicting coarse-grained semantic types of nouns and verbs (super-senses). 
\item  {\textbf{Super-sense Tagging}} ({\sc str}) As for the MWE task, we use the Streusle corpus, jointly predicting brackets and coarse-grained semantic types of the multi-word expressions. 
\end{enumerate}

\begin{table}
{
\small
\centering
\begin{tabularx}{0.48\textwidth}{lX}
\toprule
\multicolumn{2}{c}{\bf Data features}\\
\midrule
\textbf{Size}&Number of training sentences.\\
{\bf \# Labels}&The number of labels.\\
{\bf Tokens/types}&Type/token ratio in training data.\\
\textbf{OOV rate}& Percentage of training words not in GloVe vectors.\\
\textbf{Label Entropy} &Entropy of the label distribution.\\
\textbf{Frobenius norm} &$||X||_F=[\sum_{i,j} X_{i,j}^2]^{1/2}$, where $X_{i,j}$
is the frequency of term $j$ in sentence $i$.\\ 
\textbf{JSD}&Jensen-Shannon Divergence between train and test bags-of-words.\\
\midrule
\multicolumn{2}{c}{\bf Learning curve features}\\
\midrule
{\bf Curve gradients}&See text.\\
{\bf Fitted log-curve}&See text.
\end{tabularx}
\caption{\label{datafeats}Task features}
}
\end{table}

\section{Experiments}

\begin{figure}
\begin{center}
\includegraphics[scale=0.3]{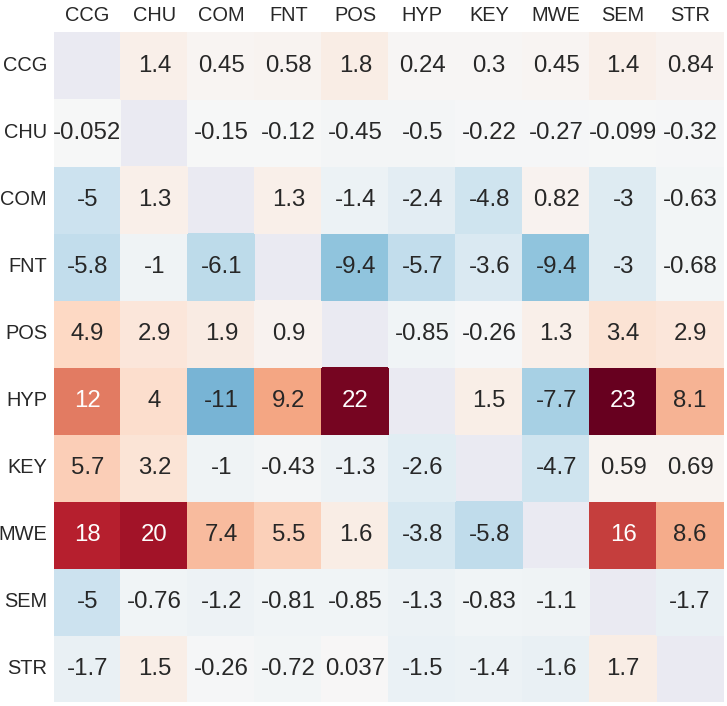} 
\caption{Relative gains and losses (in percent) over main task micro-averaged $F_1$ when incorporating auxiliary tasks (columns) compared to single-task models for the main tasks (rows).} \label{fig:heatmap}
\end{center}
\end{figure}

We train single-task bi-LSTMs for each of the ten tasks, as well as one multi-task model for each of the pairs between the tasks, yielding 90 directed pairs of the form $\langle \mathcal{T}_{main}, \{ \mathcal{T}_{main}, \mathcal{T}_{aux} \} \rangle$.
The single-task models are trained for 25,000 batches, while multi-task models are trained for 50,000 batches to account for the uniform drawing  of the two tasks at every iteration in the multi-task setup. 
The relative gains and losses from MTL over the single-task models (see Table \ref{table:datasets}) are presented in Figure \ref{fig:heatmap}, showing improvements in 40 out of 90 cases.
We see that chunking and high-level semantic tagging generally contribute most to other tasks, while hyperlinks do not significantly improve any other task. On the receiving end, we see that multiword and hyperlink detection seem to profit most from several auxiliary tasks. Symbiotic relationships are formed, e.g., by POS and CCG-tagging, or MWE and compression.

We now investigate whether we can predict gains from MTL given features of the tasks and  single-task learning characteristics. We will use the induced meta-learning for analyzing what such characteristics are predictive of gains.

Specifically, for each task considered, we extract a number of dataset-inherent features (see Table \ref{datafeats}) as well as features that we derive from the learning curve of the respective single-task model.
For the curve gradients, we compute the gradients of the loss curve at 10, 20, 30, 50 and 70 percent of the 25,000 batches. For the fitted $\log$-curve parameters, we fit a logarithmic function to the loss curve values, where the function is of the form: $L(i) = a \cdot \ln(c\cdot i+d)+b$. 
We include the fitted parameters $a$ and $c$ as features that describe the steepness of the learning curve. In total, both the main and the auxiliary task are described by 14 features. Since we also compute the main/auxiliary ratios of these values, each of our 90 data points is described by 42 features that we normalize to the $[ 0, 1]$ interval. 
We binarize the results presented in Figure \ref{fig:heatmap} and use logistic regression to predict  benefits or detriments of MTL setups based on the features computed above.\footnote{An experiment in which we tried to predict the magnitude of the losses and gains with linear regression yielded inconclusive results.} 

\begin{table}
\begin{center}

\begin{tabular}{lcc}
			& Acc. & $F_1$ (gain)\\
            \hline
Majority baseline & 0.555 & 0.615 \\
All features & 0.749 & 0.669\\
Best, data features only & 0.665 & 0.542\\
Best combination & 0.785 & 0.713\\

\end{tabular}
\caption{Mean performance across 100 runs of 5-fold CV logistic regression.} \label{table:results}
\end{center}
\end{table}

\subsection{Results} The mean performance of 100 runs of randomized five-fold cross-validation of our logistic regression model for different feature combinations is listed in Table \ref{table:results}.
The first observation is that there is a strong signal in our meta-learning features. In almost four in five cases, we can predict the outcome of the MTL experiment from the data and the single task experiments, which gives validity to our feature analysis. We also see that the features derived from the single task inductions are the most important. In fact, using only data-inherent features, the $F_1$ score of the positive class is worse than the majority baseline. 

\newcommand{\spc}{\vspace{0.7em}}
\begin{table}
\begin{center}

\begin{tabular}{lcr}
Feature	&	Task	& Coefficient \\
            \hline
Curve grad.\ (30\%)& Main & -1.566\\
Curve grad.\ (20\%)& Main & -1.164\\
Curve param. $c$   & Main &  1.007\\
\# Labels		   & Main &  0.828\\
Label Entropy	   & Aux  &  0.798\\
\spc
Curve grad.\ (30\%) & Aux&0.791\\
Curve grad.\ (50\%)&Main&0.781\\
OOV rate&Main&0.697\\
OOV rate&Main/Aux&0.678\\
Curve grad.\ (20\%)	&Aux & 0.575 \\
Fr.\ norm	&	Main & -0.516 \\ 
\spc
\# Labels	&	Main/Aux & 0.504 \\
Curve grad.\ (70\%) & Main  &	0.434 \\
Label entropy   &Main/Aux& -0.411 \\
Fr.\ norm    & Aux	& 0.346 \\
Tokens/types   &Main	& -0.297 \\
Curve param. $a$    &	Aux & -0.297 \\
\spc
Curve grad.\ (70\%)  & Aux	& -0.279 \\
Curve grad.\ (10\%)  & Aux	& 0.267 \\
Tokens/types   &Aux	& 0.254 \\
Curve param. $a$    &	Main/Aux & -0.241 \\
Size   &	Aux & 0.237 \\
Fr.\ norm    & Main/Aux	& -0.233 \\
\spc
JSD    &Aux	& -0.207 \\
\# Labels    & Aux	& -0.184 \\
Curve param. $c$    &	Aux & -0.174 \\
Tokens/types    &	Main/Aux & -0.117 \\
Curve param. $c$    &	Main/Aux & -0.104 \\
Curve grad.\ (20\%)  & Main/Aux	& 0.104 \\
\spc
Label entropy  & Main  & -0.102 \\
Curve grad.\ (50\%)  & Aux	& -0.099 \\
Curve grad.\ (50\%)  & Main/Aux	& 0.076 \\
OOV rate    & Aux	& 0.061 \\
Curve grad.\ (30\%)  & Main/Aux	& -0.060 \\
Size    & Main	& -0.032 \\
\spc
Curve param. $a$    &	Main & 0.027 \\
Curve grad.\ (10\%)  & Main/Aux	& 0.023 \\
JSD				   & Main & 0.019\\
JSD				   & Main/Aux & -0.015\\
Curve grad.\ (10\%)& Main & $6\cdot 10^{-2}$\\
Size	  		   & Main/Aux & $-6\cdot 10^{-3}$\\
Curve grad.\ (70\%)& Main/Aux & $-4\cdot 10^{-4}$\\

\end{tabular}
\caption{Strongest and weakest predictors of MTL benefit as measured by logistic regression model coefficient (absolute value).} \label{table:features}
\end{center}
\end{table}

\subsection{Analysis} Table~\ref{table:features} lists the coefficients for all 42 features. We find that features describing the learning curves for the main and auxiliary tasks are the best predictors of MTL gains. The ratios of the learning curve features seem less predictive, and the gradients around 20-30\%~seem most important, after the area where the curve typically flattens a bit (around 10\%). Interestingly, however, these gradients correlate in opposite ways for the main and auxiliary tasks. The pattern is that if the main tasks have flattening learning curves (small negative gradients) in the 20-30\%~percentile, but the auxiliary task curves are still relatively steep, MTL is more likely to work. In other words, {\em multi-task gains are more likely for target tasks that quickly plateau with non-plateauing auxiliary tasks}. We speculate the reason for this is that multi-task learning can help target tasks that get stuck early in local minima, especially if the auxiliary task does not always get stuck fast. 

Other features that are predictive include the number of labels in the main task, as well as the label entropy of the auxiliary task. The latter supports the hypothesis put forward by \newcite{Alonso:Plank:17} (see Related work). Note, however, that this may be a side effect of tasks with more uniform label distributions being easier to learn. The out-of-vocabulary rate for the target task also was predictive, which makes sense as the embedding parameters are also updated when learning from the auxiliary data. 

Less predictive features  include Jensen-Shannon divergences, which is surprising, since multi-task learning is often treated as a transfer learning algorithm \cite{Soegaard:Goldberg:16}. It is also surprising to see that size differences between the datasets are not very predictive.

\section{Conclusion and Future Work}

We present the first systematic study of when MTL works in the context of common NLP tasks, when single task parameter settings are also applied for multi-task learning. Key findings include that MTL gains are predictable from dataset characteristics and features extracted from the single-task inductions. We also show that the most predictive features relate to the single-task learning curves, suggesting that MTL, when successful, often helps target tasks out of local minima. We also observed that label entropy in the auxiliary task was also a good predictor, lending some support to the hypothesis in \newcite{Alonso:Plank:17}; but there was little evidence that dataset balance is a reliable predictor, unlike what previous work has suggested. 

In future work, we aim to extend our experiments to a setting where we optimize hyperparameters for the single- and multi-task models individually, which will give us a more reliable picture of the  effect to be expected from multi-task learning in the wild. Generally, further conclusions could be drawn from settings where the joint models do not treat the two tasks as equals, but instead give more importance to the main task, for instance through a non-uniform drawing of the task considered at each training iteration, or through an adaptation of the learning rates.  We are also interested in extending this work to additional NLP tasks, including tasks that go beyond  sequence labeling such as language modeling or sequence-to-sequence problems.

\section*{Acknowledgments}
For valuable comments, we would like to thank Dirk Hovy, Yoav Goldberg, the attendants at the second author's invited talk at the Danish Society for Statistics, as well as the anonymous reviewers.
This research was partially funded by the ERC Starting Grant LOWLANDS No.~313695, as well as by Trygfonden.
\bibliography{biblio}
\bibliographystyle{eacl2017}

\end{document}